\documentclass[conference]{IEEEtran}
\IEEEoverridecommandlockouts
\makeatletter
\let\@fnsymbol\@arabic

\usepackage{tikz}
\newcommand*\circled[1]{\tikz[baseline=(char.base)]{\node[shape=circle,draw,inner sep=2pt] (char) {#1};}}
\usepackage{amsmath}
\usetikzlibrary{shapes}
\usepackage{siunitx}
\usepackage{float}
\usepackage{hyperref}
\usepackage{cite}
\usepackage{amsmath,amssymb,amsfonts}
\usepackage{algorithmic}
\usepackage{graphicx}
\usepackage{textcomp}
\usepackage{xcolor}
\def\BibTeX{{\rm B\kern-.05em{\sc i\kern-.025em b}\kern-.08em
    T\kern-.1667em\lower.7ex\hbox{E}\kern-.125emX}}
\begin{document}

\title{Grasping Using Tactile Sensing and Deep Calibration}

\author{\IEEEauthorblockN{Masoud Baghbahari}
\IEEEauthorblockA{\textit{Central Florida Research Park} \\
\textit{University of Central Florida}\\
Orlando, USA \\
masoud@ece.ucf.edu}
\and
\IEEEauthorblockN{Aman Behal}
\IEEEauthorblockA{\textit{Central Florida Research Park} \\
\textit{University of Central Florida}\\
Orlando, USA \\
abehal@ucf.edu}
}

\maketitle

\begin{abstract}
Tactile perception is an essential ability of intelligent robots in interaction with their surrounding environments. This perception as an intermediate level acts between sensation and action and has to be defined properly to generate suitable action in response to sensed data. In this paper, we propose a feedback approach to address robot grasping task using force-torque tactile sensing. While visual perception is an essential part for gross reaching, constant utilization of this sensing modality can negatively affect the grasping process with overwhelming computation. In such case, human being utilizes tactile sensing to interact with objects. Inspired by, the proposed approach is presented and evaluated on a real robot to demonstrate the effectiveness of the suggested framework. Moreover, we utilize a deep learning framework called Deep Calibration in order to eliminate the effect of bias in the collected data from the robot sensors. 
\end{abstract}


\section{Introduction and Related Work}
In the world of active agents, an agent desires to adjust its behavior during interaction with its environment. Such an interaction consists of three main components: sensation, perception, and action. Depending on the task and the sensed modality, these three elements have to be implemented in coordination with each other to have the best performance. Perception is an intermediate step and is responsible for extracting useful information from the sensed data. The quality of extracted information from the sensed data is dependent on the power of designed perception algorithm to mine the data effectively. On the other hand, more powerful algorithms can consume a lot of time and energy from a light robot either in training phase or in inference phase. A promising solution to address these issues is integrated interactive perception approach. In such an approach, the main attention is in minimizing the perception computational processing by coordination of the three components of sensation, perception, and action. Such a coordination highlights the topic of active perception and is referred as a learned policy in the context of reinforcement learning. As is well know, learning this interactive perception from gathered experiments is computationally expensive for light robots; moreover, the learned policy can not be interpreted easily.

In this paper, we aim to propose an interactive tactile perception with use of force-torque sensing for a grasping task. Grasping task can be initiated by either visual or tactile perception. Although the first one is powerful in terms of object recognition, a continuous processing of visual data during interaction is not a simple task for a light robot. In real world, a human also barely utilizes vision ability in close proximity of the object to be grasped; indeed, tactile perception ability is more useful in the vicinity of object \cite{lee2018making}, \cite{murali2018learning}.   

Grasping is an essential and complex daily activity. Through this task, humans show an intention to affect surrounding environment in a controllable manner. Humans primarily utilize a combination of control strategy and learning from repetitive experiments to anticipate grasping in different situations \cite{romano2011human}. In this regard, the properties of an object such as size, shape, and contact surface are important parameters during grasping task \cite{lin2015robot}.

Tactile sensation is a very informative feature to recognize object properties. In \cite{chitta2011tactile}, the authors proposed a tactile perception strategy to measure tactile features for mobile robots. Tactile sensing also is used to propose a robust controller for reliable grasping \cite{eguiluz2017reliable} and slipping avoidance \cite{cirillo2017control}. Visual sensing and tactile sensing are complementary in robot grasping. A combination of both of them through deep architecture is a promising solution in \cite{guo2017hybrid}, \cite{gao2016deep}. However, processing these high-dimensional data is not an easy task and a meaningful compact representation would be needed \cite{van2016stable}, \cite{finn2016deep}. A robot can learn the manipulation using tactile sensation through demonstrations \cite{chebotar2014learning}, \cite{van2016stable}, \cite{chebotar2016self}.

A lower dimensional representation of tactile data is also more useful for object material classification \cite{kroemer2011learning}. With more processing approaches such as bag-of-words, identification of objects would be possible in advance \cite{schneider2009object}. Extraction of object pose via touch based perception can be used for manipulation  \cite{petrovskaya2007touch}. Moreover, localization will be improved by contact information gathered by tactile sensor \cite{corcoran2010measurement}, \cite{li2014localization}. The robot can control and adjust the pose of hand with stability consideration after evaluating the tactile experiences \cite{bekiroglu2011assessing}, \cite{dang2014stable}.  

Tactile feedback and interactive perception are very important components of the grasping task. An introduction of predictive force control and reactive control strategies in this domain is provided in recent years \cite{nowak2013force}. Interactive perception has been introduced as a potential field of study in recent years \cite{bohg2017interactive}. Such importance emerges from the fact that the perception can be facilitated by interaction with the environment.

In this paper, we pursue following objectives and contributions: 1) We  show that it is  feasible to propose a more human-like grasping using tactile touch sensing. 2) We describe the useful mathematical framework to extract the important tactile sensing information from robot joints. 3) We present the procedure to accurately calibrate the robot joints data in order to successfully mine the tactile sensing. 4) We propose an approach by avoiding sequential logical rules.

This paper is organized as follow. The approach on how to extract tactile sensation data is presented in Section \ref{Tactile Sensing} with grasping steps in Section \ref{Grasping Using Tactile Sensation}. Experimental results are demonstrated in Section \ref{Experimental Results} while conclusions are provided in Section \ref{Conclusions}.

\section{Tactile Sensing}\label{Tactile Sensing}
To implement our proposed approach on automatic grasping based on tactile sensing, we first describe how to extract the tactile data from the robot joint sensors data. 
\subsection{The Six-axis Force/Torque Tactile Data}
\begin{figure}
    \centering
    \begin{tikzpicture}
    \draw [inner color=blue,outer color=green, draw=black] (2.57,2.57) circle (2cm and 2cm);
    %
    \draw[thick] (4,4) -- (5,5) -- (8,2) -- (7,1);
    
    \draw[thick] (6.5,3.5) -- (7.5,4.5);
    
    \draw[red, thick,<->] (6.5,3.5) -- (6.5,3.5) node[anchor=south] {$x$};
    
    \draw[green, thick,->] (6.5,3.5) -- (8,2) node[anchor=east] {$y$};
    
    \draw[blue, thick,->] (6.5,3.5) -- (5,2) node[anchor=west] {$z$};
    
    \draw [thick,<-] (6.6,3.2) arc (-70:160:0.35cm)node[anchor=east] {$\tau_f = \tau_x$};

    \draw[thick,<-] (7.2,2.5) .. controls (8,2.75) .. (7.5,2.75) node[anchor=south] {$\tau_y$};

    \draw[black, thick,<-]  (6.2,6.2)node[anchor=south west] {$\circled{1}$} -- (4,4);
    
    \draw[black, thick,<-]  (5.5,6.5)node[anchor=south] {$\circled{2}$} -- (4,4);
    
    \draw[black, thick,<-]  (6.5,5.5)node[anchor=north] {$\circled{3}$} -- (4,4)node[anchor=south] {$f$};
    
    \draw[black, thick,-]  (8,8)node[anchor=north]{} -- (7.5,7.5)node[anchor=east] {$v$};
    
    \draw[black, thick,->]  (7.5,7.5)node[anchor=north]{} -- (7,7)node[anchor=south]{};
    
    \end{tikzpicture}
    
    \caption{Robot hand interaction with object and force/torque tactile sensing}
    \label{object and hand}
\end{figure}
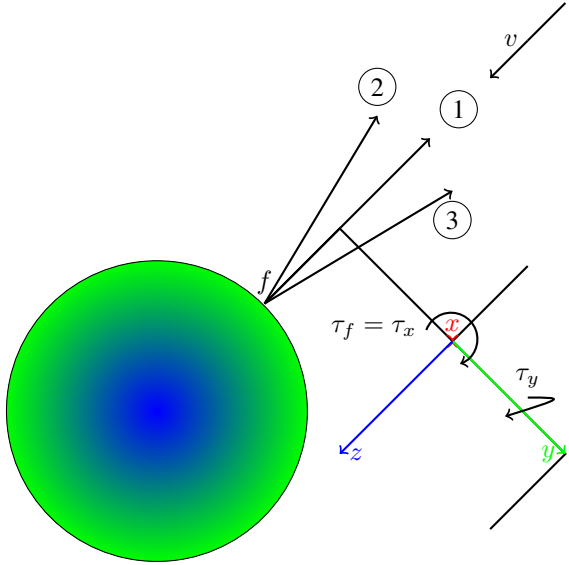

Grasping is a physical interaction with environment. During such interaction, the exchanged data between the object and robot would be the force and torque data. The robot experiences torque data $\tau_f$ as a consequence of inserted force $f$ as shown in Fig. \ref{object and hand}. 

In this figure, three directions for the sensed force are marked to be recognizable. The direction of this force ($\circled{1}$, $\circled{2}$ and $\circled{3}$) is dependent of the direction of robot hand movement toward the object $v$ and it is in opposite direction of movement after contact with the object surface:
\begin{equation}
    sign(f)=-sign(v) \label{sign_f}
\end{equation}
where the sign function returns the sign of the data. The sensed force vector can be decomposed into the robot hand frame $x$, $y$ and $z$ in Fig. \ref{object and hand}. So, the 3-axes of tactile force data are the set of $f_e=\{f_x, f_y, f_y\}$. 

The remaining part of tactile sensation is the torque data. From Fig. \ref{object and hand}, the resulting sensed torque $\tau_f$ tends to rotate the hand in the clockwise direction. This direction would be valid for all the marked force directions in the figure and since it is actually around the $x$ coordinate:
\begin{equation}
    \tau_x=\tau_f \label{equal_T}
\end{equation}
This generated torque can be expressed by cross product of the lever arm vector and the sensed force vector as follow:
\begin{equation}
    \tau_x=\Vec{r}\times\Vec{f}=||r||||f||sin\theta \label{torque_equation}
\end{equation}where $\theta$ is the angle between the lever arm vector and the force vector. 
In three dimensional space, the interaction force $f$ can also result in torque around $y$ direction denoted by $\tau_y$. The last part of torque tactile sensing set is $\tau_z$. Any movement in $x$ and $y$ coordinates during contact with object would generate a torque around $z$ axis. This torque is the consequence of friction force during movement on the object surface and the friction force is dependent on the magnitude of normal force inserted on the surface $f$. If we assume a constant friction coefficient $\mu$ for surface we can expand (\ref{torque_equation}) as:
\begin{equation}
    \tau_z=\Vec{r}\times\Vec{f}=\mu||r||||f||sin\theta \label{torque_equation_with_friction}
\end{equation}
As a consequence, the torque tactile data consists of the set of $\tau_e=\{\tau_x, \tau_y, \tau_z\}$. All the six-axis force/torque tactile data set is essential for grasping task. These set of tactile data $F_e$ are useful to guide the robot hand during interaction with the object surface:
\begin{equation}
F_e = \begin{bmatrix}
    f_e \\
    \tau_e 
    \end{bmatrix}\label{tactile_data}
\end{equation}
\subsection{Force/Torque Tactile Data and Robot Joint Sensors}
The external interaction with robot hand tends to act against the changes in Cartesian position and/or orientation via force and torque response. The force prevents the robot from further changes in the Cartesian position whilst the torque is against the change in orientation. Such external effects can sensed as torque in each joint of the robot. The relationship between the robot measured joint torque sensors and the interaction force torque in robot base frame (the frame attached to the first joint holding the whole robot arm) can be stated by the transpose of the Jacobian matrix $J$: 
\begin{equation}
\tau_{int} = J^TF \label{eq3}
\end{equation}
where $\tau_{int}$ is the interaction torque data sensed by robot joint torque sensors and $F$ is the vector of six elements encapsulating the end-effector interaction force $f_b$ and torque $\tau_b$ expressed in the base frame:
\begin{equation}
F = \begin{bmatrix}
    f_b \\
    \tau_b 
    \end{bmatrix}\label{base_force}
\end{equation}
Assuming a Jacobian matrix with number of rows equal or greater than 6, the interaction force-torque $F$ on end-effector can be retrieved by the Moore-Penrose inverse of the Jacobian matrix:
\begin{equation}
    F = (JJ^T)^{-1}J\tau_{int} \label{Pseudo_inverse}
\end{equation}
Since the Jacobian is a joint dependent matrix, the inverse term in some specific joint positions known as singularities is not defined. In our case, we assume that our robot with 6 joints never works in such configurations of the robot arm. 
Although the force $f_b$ and the torque $\tau_b$ magnitude are independent of the frame, their directions are completely dependent on the expressed frame. As shown in Figure \ref{Coordinate_trasnformation}, transformation of these data from base frame $(b)$ to end-effector frame $(e)$ is possible using the relative Rotation Matrix $R_b^e$:
\begin{figure}
    \centering
    \begin{tikzpicture}

        \draw [->] (0,0,0) -- (2,0,0) node [right] {$y_b$};
        \draw [->] (0,0,0) -- (0,2,0) node [left] {$z_b$};
        \draw [->] (0,0,0) -- (0,0,2) node [left] {$x_b$};
    
        \draw [thick,->] (0.55,0.45,0.45) parabola (3.55,3.45,3.45)  node[anchor = north west] {$R_b^e$};
    
        \draw [thick,->] (0.55,0.45,0.45) parabola (3.55,3.45,3.45)  node[anchor = north west] {$R_b^e$};
    
        \draw [->] (3,3,3) -- (2,3,3) node [left] {$y_e$};
        \draw [->] (3,3,3) -- (3,4,3) node [left] {$x_e$};
        \draw [->] (3,3,3) -- (3,3,4) node [left] {$z_e$};
        
        \draw [thick,->] (0,0,0) -- (1,1,1)node [right]{$vec$};
    
        \draw [thick,->] (3,3,3) -- (4,4,4)node [right]{$vec$};
    
    \end{tikzpicture}

    \caption{A typical vector $vec$ coordinate transformation by rotation matrix $R_b^e$. The magnitude of vector is same in both coordinates.}
    \label{Coordinate_trasnformation}
\end{figure}
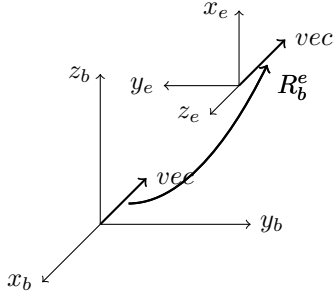
\begin{equation}
\begin{bmatrix}
    f_e \\
    \tau_e 
    \end{bmatrix} =
    \begin{bmatrix}
    R_b^e & 0 \\
    0  & R_b^e
    \end{bmatrix}\begin{bmatrix}
    f_b \\
    \tau_b 
    \end{bmatrix}\label{force_trasnform}
\end{equation}
where $f_e$ and $\tau_e$  are the force and the torque in end-effector frame obtained as force torque tactile data in (\ref{tactile_data}) as shown in Fig. \ref{object and hand}. Using both (\ref{Pseudo_inverse}) and (\ref{force_trasnform}), it is possible to retrieve the six-axis force/toque tactile data from the joint sensed interaction torques data $\tau_{int}$.

\subsection{Deep Calibration: Robot Joint Sensors Calibration Using Deep Learning}
In light of (\ref{Pseudo_inverse}), we assume that the robot joint sensors data $\tau_s$ is purely related to the interaction with object. However, this data needs extra processing of calibration to accurately calculate the tactile data. In this section, we propose to utilize a deep learning technique to calibrate the joint torque sensors. These torques capture many effects, related to the robot motion and gravity as well as the interaction with the object. As a consequence, this bias is required to be removed to extract the interaction tactile data successfully. 

To remove the motion-related torque data, we proposed to model the motion bias torque data. The interaction bias-free joint torque data will be obtained by subtracting the measured joint torque data $\tau_s$ from this bias model $\tau_{bias}$:
\begin{equation}
    \tau_{int} = \tau_s - \tau_{bias}\label{bias_error}
\end{equation}

In this regard, we record the robot joints sensor torque data as well as the joints position $q$ and velocity $\dot{q}$ during robot free-motion in space. Our primary experiments demonstrate that the motion-related torque data change considerably with two parameters: joint movement direction $\dfrac{\dot{q}}{|\dot{q}|}$ and the joint angle $q$. These two variables are fed as inputs to a three-layer fully-connected sequential multilayer perceptron deep network to train and inference the bias value from the output of the model. In fact, the model captures the significant changes in the collected bias data through fitting a regression to the bias data \cite{baghbahari2018evaluation}. 
\section{Grasping Using Tactile Sensing}\label{Grasping Using Tactile Sensation}
In this section, we present the procedure to define the grasping steps for a robot using the six-axis tactile data acquired in the previous section. 
\subsection{Feedback Control Command}
In this subsection, we present the procedure to define the grasping steps for a robot using the six-axis tactile data in previous section. We can use the interaction force/torque data to guide the robot hand in the vicinity of the object. These data are informative enough to define the required grasping steps. Our approach is based on feedback tactile sensation to adjust the robot hand velocity in corresponding direction during its interaction with the object (see Fig. 
\ref{object and hand}). With this goal, we assume dynamics for linear velocity $v$ and angular velocity $\omega$ of robot hand as follows \cite{baghbahari2018real}:
\begin{equation}
    \dot{v} + b_vv = u_v\label{dynamic_linear}
\end{equation}
\begin{equation}
    \dot{\omega} + b_{\omega}\omega = u_\omega \label{dynamic_angular}
\end{equation}
where $u_v$ is the command to adjust the linear velocity and $u_\omega$ to adjust the angular velocity in which the direction of linear and angular velocities are expressed in the robot hand frame itself (Fig. \ref{object and hand}). The constant coefficients $b_v$ and $b_{\omega}$ are the system damping to inertia ratio and determine the time profile of linear and angular velocity in response to the adjustment command respectively. 

According to Fig. \ref{object and hand}, any change in hand location around the object surface will be effected by commanding the linear velocity. Meanwhile, any rotation around the object to point the hand fingers toward the object can be accomplished via influencing the angular velocity. As the robot fingers are interacting with the object surface, the adjustment in the commands is needed to change the hand location or the angle between the hand and the object. Our approach is based on implementing the tactile sensation to adjust the hand linear and angular velocity during its interaction with the object. The control command for linear velocity is defined as:
\begin{equation}
    u_v=\alpha_v(f_f-f_e)\label{force_feedback}
\end{equation}
where $f_e$ can be any element of force tactile sensation set $f_e=\{f_x, f_y, f_z\}$ and $\alpha_v$ is a necessary scaling constant to scale two different domains (force in several $Newton$ and velocity typically in less than several $mm/s$). The desired value $f_f$ is the desired final value that we consider for that direction. For instance, if the robot needs to touch the object, it has to move directly toward the object (see Fig. \ref{object and hand}). In such case, the control command has to be designed for linear velocity in hand $z$ direction with $\alpha_{vz}<1$:
\begin{equation}
    u_{vz}=\alpha_{vz}(f_{zf}-f_z)\label{force_approaching}
\end{equation}
Executing (\ref{dynamic_linear}) with this command, the robot hand moves with a constant velocity $v_{dz}$ given as
\begin{equation}
    v_{dz}=\dfrac{\alpha_{vz}f_{zf}}{b_{z}}\label{force_approaching}
\end{equation}
toward the object until the force tactile sensation $f_z$ converges to $f_{zf}$. Nevertheless, the velocity of movement is dependent on the final touch force $f_{dz}$. With a small modification, we replace the control command with following alternative:
\begin{equation}
    u_{vz}=b_v v_{dz}(1 + \alpha_{vz} f_{z})\label{force_approaching_modify}
\end{equation}
where $v_{dz}$ can be selected freely and the final contact force is adjusted by $\alpha_{vz}$ as follows:
\begin{equation}
    f_{zf}=-\dfrac{1}{\alpha_{vz}}.\label{touch_force}
\end{equation}
Similarly, to rotate the hand around each axis, the corresponding control command can be defined by a scaling coefficient
\begin{equation}
    u_\omega=\alpha_{\omega}(\tau_d-\tau_e)\label{torque_feedback}
\end{equation}
where $\tau_e=\{\tau_x, \tau_y, \tau_z\}$ is the feedback torque tactile sensation. As another example, if we need to align the hand by rotating the hand around its $z$ axis, we can consider:
\begin{equation}
    u_{\omega z}=\alpha_{\omega z}(\tau_{dz}-\tau_z)\label{torque_z}
\end{equation}
Inserting this command into (\ref{dynamic_angular}) after a transient time imposed by the system damping to inertia ratio in the direction $b_{\omega z}$, the rotation with a constant angular velocity $\omega_{z}$ equals
\begin{equation}
    \omega_z=\dfrac{\alpha_{\omega z}\tau_{dz}}{b_{\omega z}} \label{desired_angular_z}
\end{equation}
which will continue till the corresponding tactile sensing measurement $\tau_z$ reaches the level of desired torque $\tau_{dz}$. We also can define a desired value as a function of other sensed data. As an example, if the robot needs to rotate around the $z$ direction as a function of force in $z$ direction, (\ref{torque_z}) changes to:
\begin{equation}
    u_{\omega z}=\alpha_{\omega z}(\beta_{\omega z}*f_{z}-\tau_z)\label{torque_z_modify}
\end{equation}
where $\beta_{\omega z}$ is scaling factor and the rotation is effected until the desired force in $z$ direction converges to the desired value $f_{dz}$.
If the desired value is related to other tactile sensing data, the adjustments would be coupled to one another. This is a very important property since we can define a sequence of actions for the grasping task. We define an intuitive grasping paradigm according to Fig. \ref{object and hand} as follows:
\begin{itemize}
    \item Move directly toward the object in $z$ direction
    \item Stop movement when the fingers touch the object
    \item Keep a constant inserted force in the $z$ direction
    \item Rotate around the contact point using the torque tactile data
\end{itemize}

All these defined steps can be executed by both proposed commands in (\ref{force_approaching_modify}) and (\ref{torque_z}) with some minor modifications. The three first steps are encapsulated in (\ref{force_approaching_modify}). To make sure that the inserted force is in $z$ direction, we need to modify (\ref{force_approaching_modify}) as follow for both $x$ and $y$ directions:
\begin{equation}
    u_{xz}= \alpha_{vx} f_{x}\label{force_x}
\end{equation}
\begin{equation}
    u_{yz}= \alpha_{vy} f_{y}\label{force_y}
\end{equation}
where $\alpha_{vx}$ and $\alpha_{vy}$ are scaling coefficients in those directions. Under the implementation of these two extra commands, the robot always inserts the force in $z$ direction. Finger elasticity is the main reason for sensing force in $x$ and $y$ directions. Such forces can be eliminated easily to help the robot contact with the object in just the $z$ direction. As a result, we can plan for $x$ and $y$ directions to facilitated the grasping in advance settings.

To rotate around the contact point using tactile torque data, we propose the following commands:
\begin{equation}
    u_{\omega x}=\alpha_{\omega x}(\tau_x)\label{torque_x}
\end{equation}
\begin{equation}
    u_{\omega y}=\alpha_{\omega y}(\tau_y)\label{torque_y}
\end{equation}
where $\alpha_{\omega x}$ and $\alpha_{\omega y}$ are the scaling factors for the corresponding directions. The sensed torques facilitate alignment of each direction with the object surface at the end of the rotation.

In the next section, we prove the stability of above dynamic equations with designed control commands.
\subsection{Stability Analysis}
Since the proposed approach is based on feedback data, we need to prove the stability of the real-time dynamic system under implementation of the proposed control commands.
We first investigate the stability of (\ref{dynamic_linear}) in $z$ direction with  (\ref{force_approaching_modify}):
\begin{equation}
    \dot v_z + b_z v_z = b_z v_{dz} (1 + \alpha_{vz} f_z)  \label{proof_force_z}
\end{equation}
Before interaction, the force sensing data is zero and the system follows the following dynamic equation to move the robot toward the object:
\begin{equation}
    \dot v_z + b_z (v_z - v_{dz})=0  \label{eq24}
\end{equation}
This is a stable dynamic equation and the error converges to zero after a settling time greater than $t_s = \dfrac{4}{b_z}$:
\begin{equation}
    (v_z - v_{dz}) \,\to\, 0  \label{eq25}
\end{equation}
which implies $v_z \,\to\, v_{dz}$ when the sensed force $f_z = 0$.
After contact with the object, the sensed force is no longer zero. Due to flexibility of fingers, the sensed force can be modeled as a spring force with coefficient $K_z$. It is proportionally dependent on the level of displacement in the finger joints as well as the object deformation expressed into a unified term $\Delta z$ in $z$ direction:
\begin{equation}
    f_z = -K_z\Delta z \label{force_model_z}
\end{equation}
We also can model the unified term $\Delta z$ as the displacement from equilibrium point of the surface (see Fig. \ref{displacement}) and insert it into (\ref{proof_force_z}):
\begin{equation}
    \dot v_z + b_z v_z = b_z v_{dz} (1 - \alpha_{vz} K_z\Delta z)  \label{eq26}
\end{equation}
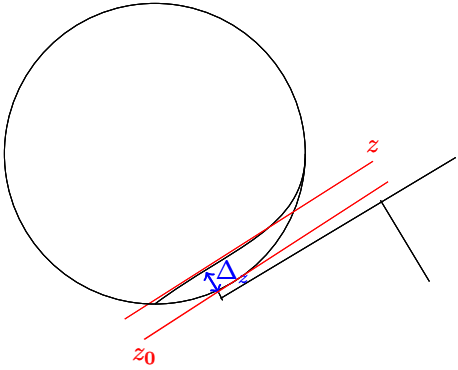
\begin{figure}[ht]
    \centering
    \begin{tikzpicture}

    \draw (2,2) arc (0:-90:2cm);
    \draw (0,0) .. controls (1,0.75) and (2,1) .. (2,2);
    \draw (0.9,0.05) -- (0.85,0.15);
    \draw (0.875,0.075) -- (4,1.95);
    \draw (3,1.38) -- (3.65,0.3);
    
    \draw (2,2) arc (0:-90:2cm);
    \draw (0,0) .. controls (1,0.75) and (2,1) .. (2,2);
    \draw (0.9,0.05) -- (0.85,0.15);
    \draw (0.875,0.075) -- (4,1.95);
    \draw (3,1.38) -- (3.65,0.3);
    
    \draw (2,2) arc (0:-90:2cm);
    \draw (0,0) .. controls (1,0.75) and (2,1) .. (2,2);
    \draw (0.9,0.05) -- (0.85,0.15);
    \draw (0.875,0.075) -- (4,1.95);
    \draw (3,1.38) -- (3.65,0.3);

    \draw[blue, thick,<->] (0.85,0.15) -- (0.68,0.425) node[anchor=west] {$\Delta_z$};
    \draw[blue, thick,<->] (0.85,0.15) -- (0.68,0.425) node[anchor=west] {$\Delta_z$};


    \draw [red](-0.4,-0.2) -- (5*0.58,5*0.38) node[anchor=south] {$z$};
    
    \draw [red](-0.4,-0.2) -- (5*0.58,5*0.38) node[anchor=south] {$z$};
    
    \draw[red] (0.775*4,0.775*2.1) -- (-0.145,-2*0.235) node[anchor=north] {$z_0$};
    
    \draw[red] (0.775*4,0.775*2.1) -- (-0.135,-2*0.235) node[anchor=north] {$z_0$};

    
    \draw (2,2) arc (0:270:2cm);
    \draw (2,2) arc (0:270:2cm);
    \draw (2,2) arc (0:270:2cm);

    \end{tikzpicture}
    \caption{Displacement model springiness term during interaction with object} \label{displacement}
\end{figure}
\begin{equation}
    \Delta z = z-z_0  \label{eq27}
\end{equation}
where $z$ is the location of finger tip and $z_0$ is the equilibrium point of surface, both of them in the $z$ direction. Rearranging (\ref{eq26}), we have the following:
\begin{equation}
    \dot v_z + b_z v_z + \alpha_{vz} K_z\Delta z= b_z v_{dz}. \label{eq28}
\end{equation}
Moreover, we know that the velocity is the rate of change in location
\begin{equation}
    v_z = \dot z \label{eq29}
\end{equation}
such that we we can modify (\ref{eq28}):
\begin{equation}
    \ddot z + b_z  \dot z + b_z v_{dz} \alpha_{vz} K_z (z - z_0)= b_z v_{dz} \label{eq30}
\end{equation}
This is a second order stable dynamic equation. The velocity error and the acceleration error become zero after a settling time around $t_s = \dfrac{4}{b_z}$ and the sensed force converges to
\begin{equation}
    f_z = -K_z (z_d - z_0) = -\dfrac{b_z v_{dz}}{b_z v_{dz} \alpha_{vz}}=-\dfrac{1}{\alpha_{vz}}
    \label{eq31}
\end{equation}
where $z_d$ is the final value when the robot hand finger can not move any further. This proof can be easily extended to other directions.
The results are extendable for sensed torque as well. In such a case, we utilize the mathematical relationship for sensed torque when there is an angle $\theta$ between the force vector $\vec{f}$ and lever arm vector $\vec{d}$:
\begin{equation}
    \vec{\tau} = \vec{d}\times \vec{f} = |d||f|sin(\theta). \label{eq35}
\end{equation}
For rotation around the hand $z$ direction, the sensed force $\vec{f}$ is perpendicular to the lever arm vector $\vec{d}$. With a friction coefficient $\mu_z$ during rotation on the surface, the sensed torque would be: 
\begin{equation}
    \tau_z = d \mu_z f_z sin(\ang{90}) = d \mu_z f_z \label{torque_z_const}
\end{equation}
After utilizing this equation in (\ref{torque_z}) and inserting the sensed torque into (\ref{dynamic_angular}), we have
\begin{equation}
   \dot \omega_{z} + b_{\omega z} \omega_{z} = \alpha_{\omega z}(-d \mu_z f_z + \beta_{\omega z}f_z) = \alpha_{\omega z}f_z(-d\mu_z+\beta_{\omega z}). \label{dynamic_angular_z}
\end{equation}
According to this equation, rotation around the $z$ direction will be effected as long as the sensed force $f_z \neq 0$. The rotation by commanding the angular velocity $\omega_z$ continues on the surface until it detects the edges. In this case, the robot stops rotating around the $z$-axis when the following condition is satisfied
\begin{equation}
   \beta_{\omega z} = d\mu_z. \label{eq37}
\end{equation}
If we select $\beta$ bigger than $d\mu_z$, the robot only reaches the edges where $\mu_z$ is bigger than the surface friction coefficient. In this case, the sensed torque converges to:
\begin{equation}
   \tau_z = \beta_{\omega z} f_{zf} \label{eq37}
\end{equation}
when the object is inside the robot hand and $f_z \,\to\, f_{zf}$.
\section{Experimental Results}\label{Experimental Results}
In this section, we present the experimental settings, the time profile of tactile sensation set for a typical grasping, and the successful rate for a set of different objects.
\subsection{Settings} 
To validate the proposed approach, we use a two fingered Mico robot with 6 degrees of freedom. Except for singular configurations and some specific configurations in which the robot is unable to reach, the Jacobian matrix used in (\ref{Pseudo_inverse}) results in valid tactile sensation data. The deep learning is trained with ReLU activation function for the neurons during 50 epochs and a batch size of 20 to minimize the mean square error for each joint torque bias data. The configuration of final layer is the output estimating the bias value corresponding to the current joint angle and movement direction.

The sensor data is collected through robot movement without any interaction with external environment. By this way, the gathered data is only related to the current joint angle and the direction of the movement completely reflecting the dependency of bias torque data to these inputs. 
\subsection{Grasping Paradigm Using Tactile Sensing}
\begin{figure}[h!]
      \centering
      \parbox{3.5in}{\includegraphics[scale=0.6]{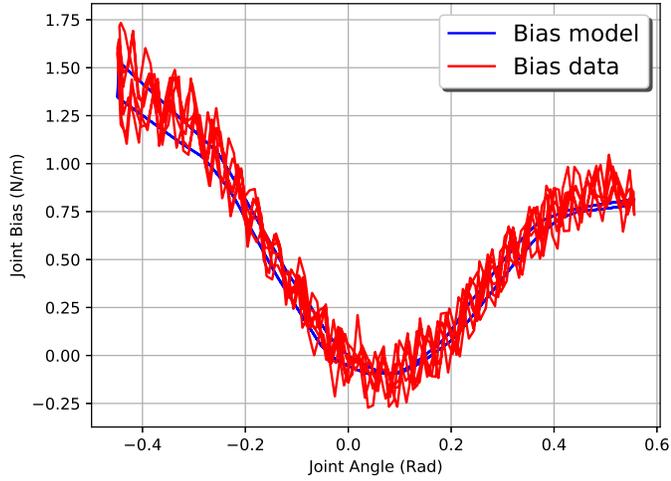}}
      \caption{Bias data and fitted model of first joint}
      \label{joint1_bias}
\end{figure}

\begin{figure}[h!]
      \centering
      \parbox{3.5in}{\includegraphics[scale=0.6]{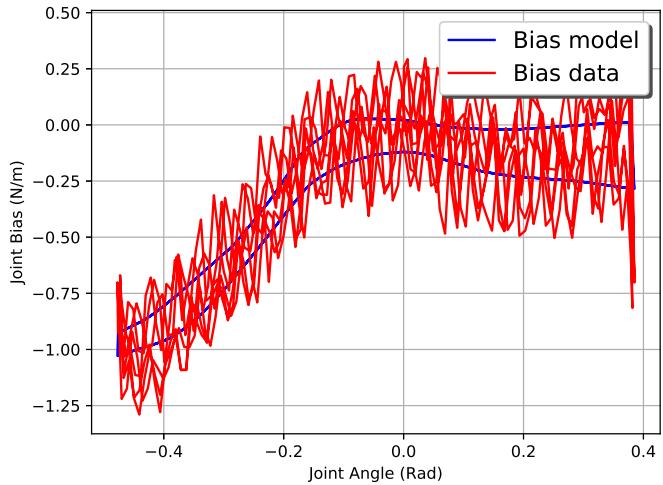}}
      \caption{Bias data and fitted model of second joint}
      \label{joint2_bias}
\end{figure}

\begin{figure}[h!]
      \centering
      \parbox{3.5in}{\includegraphics[scale=0.6]{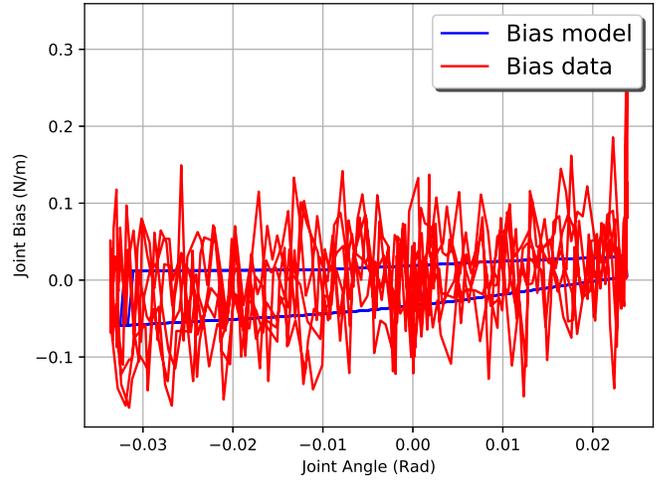}}
      \caption{Bias data and fitted model of third joint}
      \label{joint3_bias}
\end{figure}

\begin{figure}[h!]
      \centering
      \parbox{3.5in}{\includegraphics[scale=0.6]{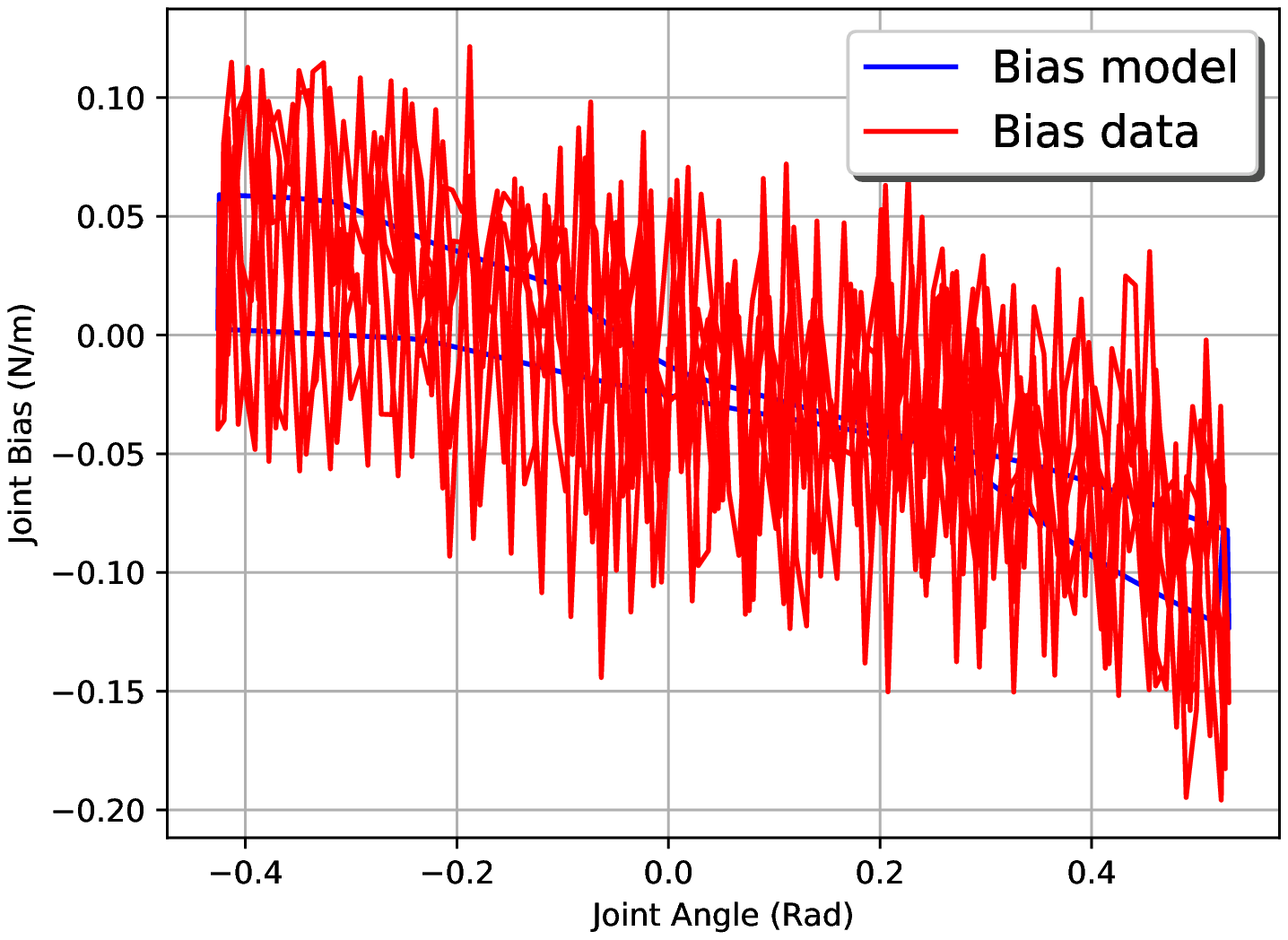}}
      \caption{Bias data and fitted model of forth joint}
      \label{joint4_bias}
\end{figure}

\begin{figure}[H]
      \centering
      \parbox{3.5in}{\includegraphics[scale=0.6]{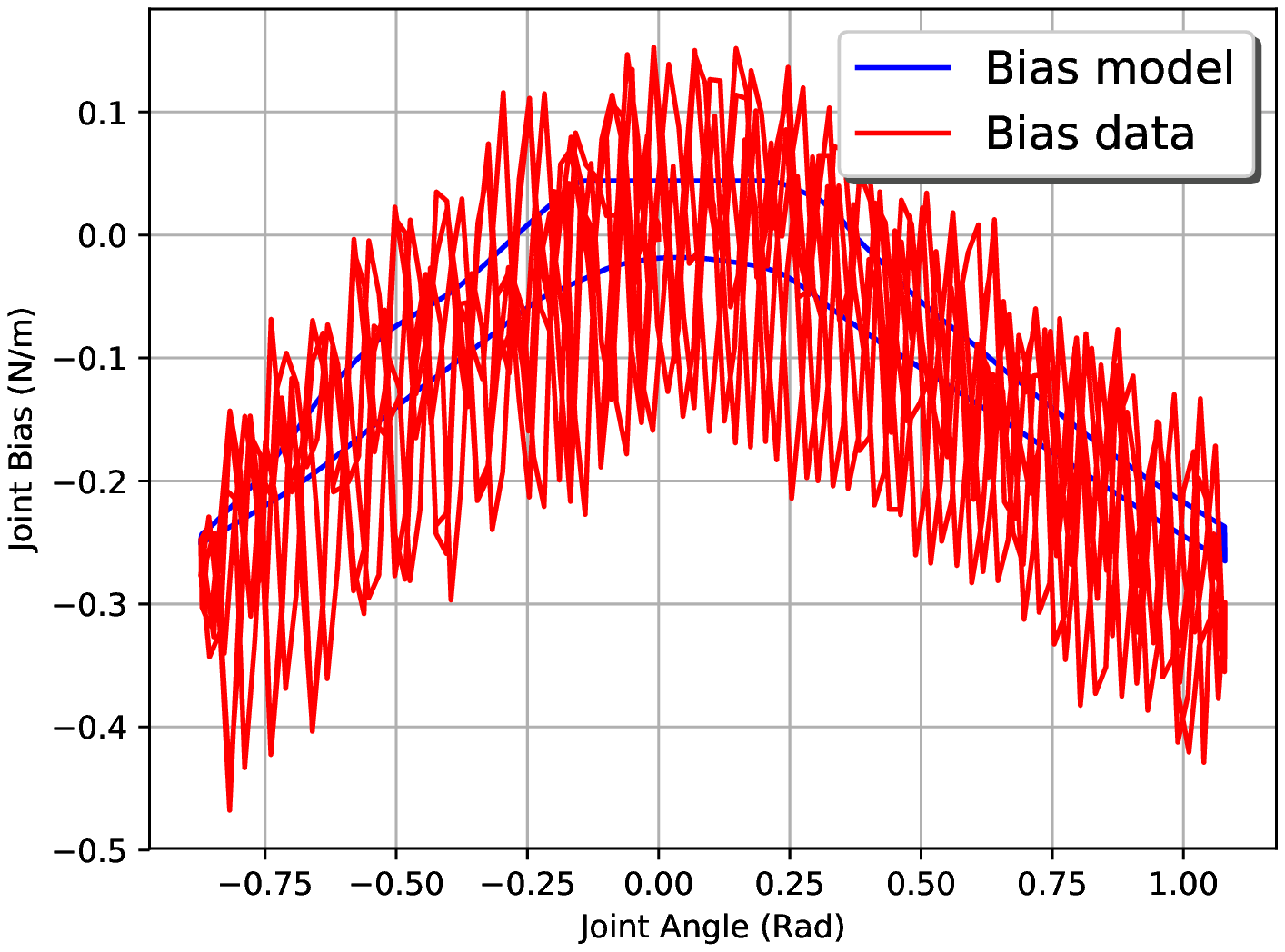}}
      \caption{Bias data and fitted model of fifth joint}
      \label{joint5_bias}
\end{figure}

\begin{figure}[H]
      \centering
      \parbox{3.5in}{\includegraphics[scale=0.6]{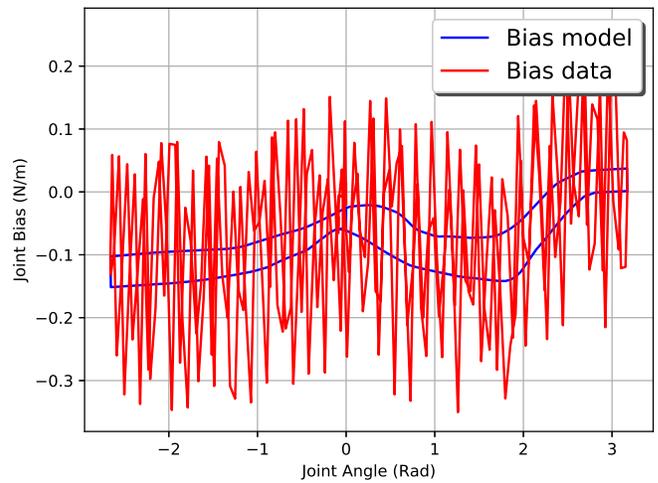}}
      \caption{Bias data and fitted model of sixth joint}
      \label{joint6_bias}
\end{figure}

We first evaluate the deep calibration performance. In Fig. \ref{joint1_bias} through Fig. \ref{joint6_bias}, the recorded torque data and the model are visualized for all six joints to illustrate the performance of proposed deep calibration in details. There are two blue lines on the vibrating noisy data and each of them is related to the direction of joint movement. In prediction phase, the torque data is extracted from the network corresponding to the joint angle and its movement direction. 

\begin{figure}[H]
      \centering
      \parbox{3.5in}{\includegraphics[scale=0.6]{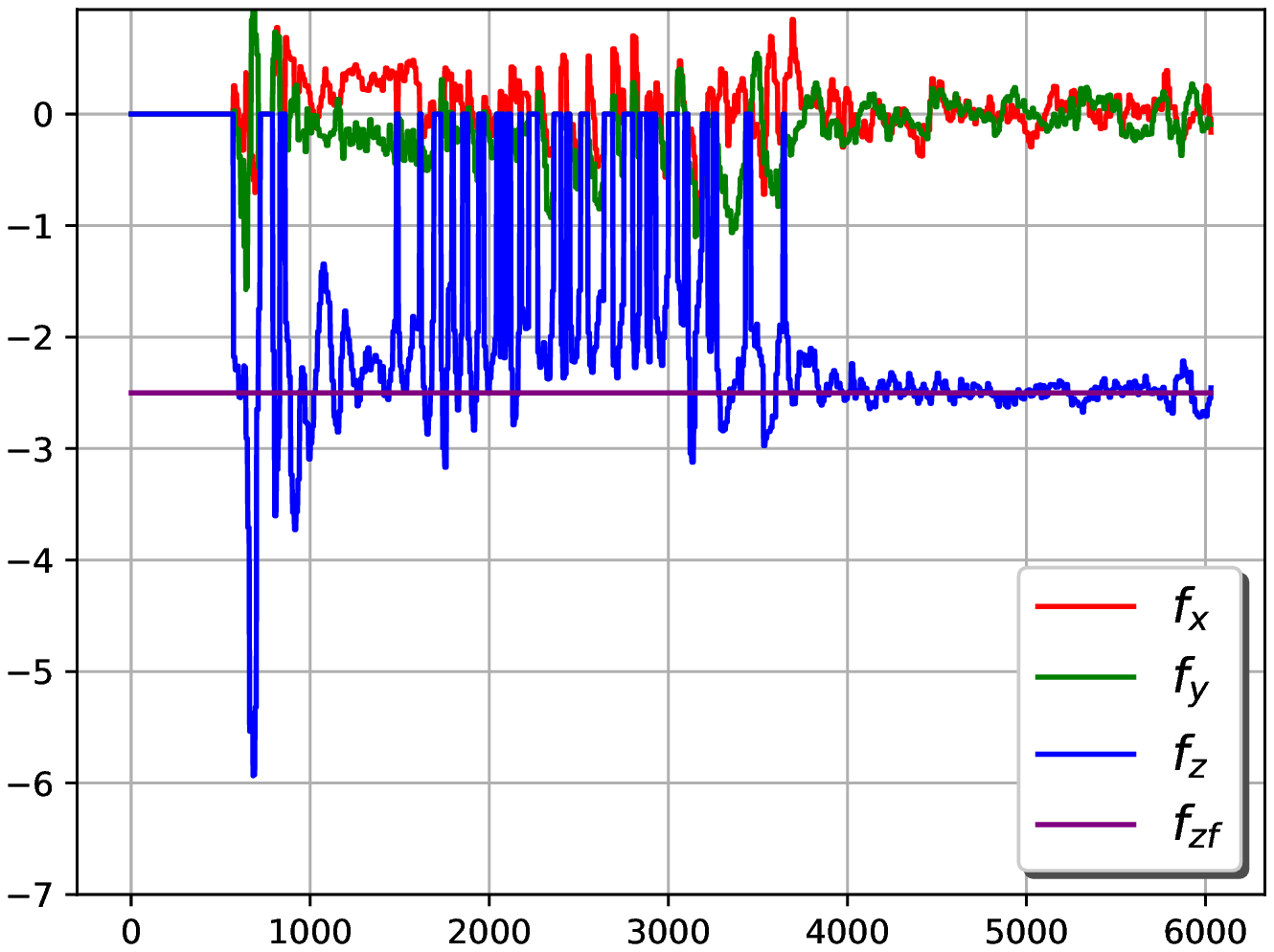}}
      \caption{Force tactile sensation profile during grasping the canned}
      \label{force_tactile_sensing}
\end{figure}

Next, we apply the grasping paradigm in this paper to grasp a canned. In Table \ref{table:parameters}, we provide parameters involved in the design. Using these parameters, the commands executed in the grasping process are as follows:
\begin{equation}
    u_{xz}= 0.0025 f_{x}\label{exp:force_x}
\end{equation}
\begin{equation}
    u_{yz}= 0.0025 f_{y}\label{exp:force_y}
\end{equation}
\begin{equation}
    u_{vz}= v_{dz}(1 + 0.4 f_{z})\label{exp:force_z}
\end{equation}
\begin{equation}
    u_{\omega x}= (0.0025*\tau_x)\label{exp:torque_x}
\end{equation}
\begin{equation}
    u_{\omega y}= (0.0025*\tau_y)\label{exp:torque_y}
\end{equation}
\begin{equation}
    u_{\omega z}= (0.025*f_{z}-\tau_z)\label{exp:torque_z}
\end{equation}
According to these commands, the tactile force sensing in approaching direction $z$ would be
\begin{equation}
    f_{z}=-\dfrac{1}{\alpha_{vz}} = -\dfrac{1}{0.4}=-2.5 N \label{exp:final_force_z}
\end{equation}
which is consistent with the final value of force tactile sensing in Fig. \ref{force_tactile_sensing}. Moreover, the final values of force sensing in $x$ and $y$ directions are zero. As shown in the figure, the robot reaches the object surface after 6000 time samples which is equal to $6$ $sec$ since the sample time is $0.001$ $sec$. Nevertheless, the grasping task is completed after $35$ $sec$ of touching the object surface since the robot struggles to adjust the hand with the object surface before grasping the object completely. It is hard for the robot to keep a constant contact with the object while is suffering from mechanical vibration during movement. Several jumps in Fig. \ref{force_tactile_sensing} demonstrate this phenomenon. 

\begin{table}[H]
    \caption{Parameters}
    \label{table:parameters}
    \begin{center}
    \begin{tabular}{|c||c|}
    \hline
    Parameter & Value\\
    \hline
    $v_{dz}$ & 0.0055\\
    \hline
    $\alpha_{vx}$ & 0.0025\\
    \hline
    $\alpha_{vy}$ & 0.0025\\
    \hline
    $\alpha_{vz}$ & 0.4\\
    \hline
    $\alpha_{\omega x}$ & 0.25\\
    \hline
    $\alpha_{\omega y}$  & 0.0025\\
    \hline
    $\alpha_{\omega z}$  & 1\\
    \hline
    $\beta{\omega z}$ & 0.025\\
    \hline
    $b_x$  & 1\\
    \hline
    $b_y$  & 1\\
    \hline
    $b_z$  & 1\\
    \hline
    $b_{\omega x}$  & 1\\
    \hline
    $b_{\omega y}$  & 1\\
    \hline
    $b_{\omega z}$  & 1\\
    \hline
    \end{tabular}
    \end{center}
\end{table}
\begin{figure}[H]
      \centering
      \includegraphics[scale=0.6]{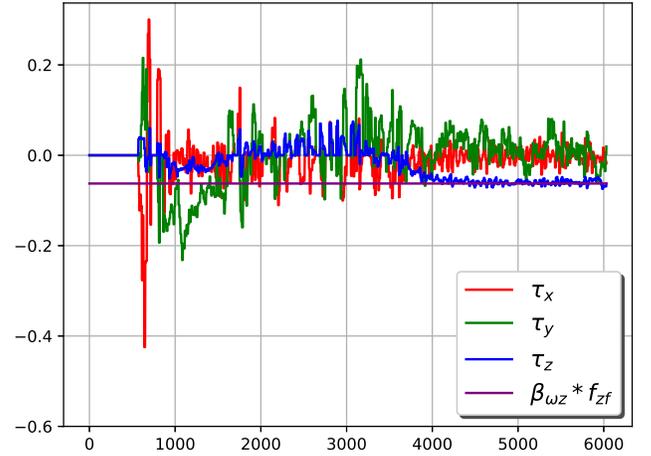}
      \caption{Torque tactile sensation profile during grasping the canned}
      \label{torque_tactile_sensing}
\end{figure}

During this typical interaction, the torque tactile sensing reaches zero for both $x$ and $y$ directions and it converges to a constant final value:
\begin{equation}
    \tau_{zf} = \beta_{\omega z}*f_{zf} = -0.025*0.25 = -0.0625
\end{equation}
in the $z$ direction when it detects the object edges. All these explanations can be observed in Fig. \ref{torque_tactile_sensing}.

As can be inferred from these figures, the sensed data is prone to mechanical vibrations during the robot movement as well as noise. We applied a threshold filter on the data to reduce this undesirable effect after removing the bias using deep calibration. This is the reason that the sensed data is zero before contact with the object. This filter has no effect on the data vibration after it passes the threshold. According to presented results, our proposed approach grasp an unknown object similar to human being. Deep calibration is useful to provide the accurate tactile sensing data. This data can be mined from robot joint sensors by a mathematical formulation and the grasping strategy implemented by feedback tactile data works smoothly without implementation any logical rules. A video of the experiment is available online \url{https://youtu.be/y7ZBFr1IpVw}.
\section{Conclusions}\label{Conclusions}
In this paper, we proposed a human-like automatic grasping of an object using tactile sensing without vision. The tactile sensing data is extracted from the robot joint torque sensors and calibration process is needed to rid the tactile data of the bias related to motion. To calibrate the bias data, we provided a customized deep learning structure with effective inputs to fit a model on bias data. The suggested grasping paradigm in this paper is applied to grasp an unknown shaped canned and the time profile of tactile sensing data is presented during execution of grasping with deep learning alongside the feedback tactile data.

As part of future works, our goal is to propose enhancement strategies for manipulation capability of robots using tactile sensing. In this regard, harder manipulation tasks will be addressed through a touch-based interactive  perception.

\bibliographystyle{unsrt}  

\bibliography{references}

\end{document}